\documentclass{article}
\usepackage{arxiv}
\usepackage[utf8]{inputenc} % allow utf-8 input
\usepackage[T1]{fontenc}    % use 8-bit T1 fonts
\usepackage{hyperref}       % hyperlinks
\usepackage{url}            % simple URL typesetting
\usepackage{booktabs}       % professional-quality tables
\usepackage{amsfonts}       % blackboard math symbols
\usepackage{nicefrac}       % compact symbols for 1/2, etc.
\usepackage{microtype}      % microtypography
\usepackage{lipsum}		% Can be removed after putting your text content
\usepackage{graphicx}
\usepackage{natbib}
\usepackage{doi}
\usepackage{color}
\usepackage{xcolor}
\usepackage{colortbl}
\usepackage{bbding}
\usepackage{multirow}
\usepackage{pifont}

\title{Mobile-Agent: Autonomous Multi-Modal Mobile Device Agent with Visual Perception}

\author{Junyang Wang$^1$\thanks{Work done during internship at Alibaba Group.} \quad Haiyang Xu$^2$\thanks{Corresponding author} \quad Jiabo Ye$^2$ \quad Ming Yan$^2$\footnotemark[2] \and \textbf{Weizhou Shen$^2$ \quad Ji Zhang$^2$ \quad Fei Huang$^2$ \quad Jitao Sang$^1$\footnotemark[2]}\\ \\
{\tt \{junyangwang, jtsang\}@bjtu.edu.cn, \{shuofeng.xhy, ym119608\}@alibaba-inc.com}\\ \\
$^1$Beijing Jiaotong University \quad $^2$Alibaba Group
}

\hypersetup{
pdftitle={A template for the arxiv style},
pdfsubject={q-bio.NC, q-bio.QM},
pdfauthor={David S.~Hippocampus, Elias D.~Striatum},
pdfkeywords={First keyword, Second keyword, More},
}

\begin{document}

\definecolor{darkgreen}{rgb}{0.0, 0.5, 0.0}

\maketitle

\begin{abstract}

Mobile device agent based on Multimodal Large Language Models (MLLM) is becoming a popular application. In this paper, we introduce Mobile-Agent, an autonomous multi-modal mobile device agent. Mobile-Agent first leverages visual perception tools to accurately identify and locate both the visual and textual elements within the app's front-end interface. Based on the perceived vision context, it then autonomously plans and decomposes the complex operation task, and navigates the mobile Apps through operations step by step. Different from previous solutions that rely on XML files of Apps or mobile system metadata, Mobile-Agent allows for greater adaptability across diverse mobile operating environments in a vision-centric way, thereby eliminating the necessity for system-specific customizations. To assess the performance of Mobile-Agent, we introduced Mobile-Eval, a benchmark for evaluating mobile device operations. Based on Mobile-Eval, we conducted a comprehensive evaluation of Mobile-Agent. The experimental results indicate that Mobile-Agent achieved remarkable accuracy and completion rates. Even with challenging instructions, such as multi-app operations, Mobile-Agent can still complete the requirements. Code and model are open-sourced at \url{https://github.com/X-PLUG/MobileAgent}.

\end{abstract}

\begin{figure*}[!ht]
    \centering
    \includegraphics[width=0.6\textwidth]{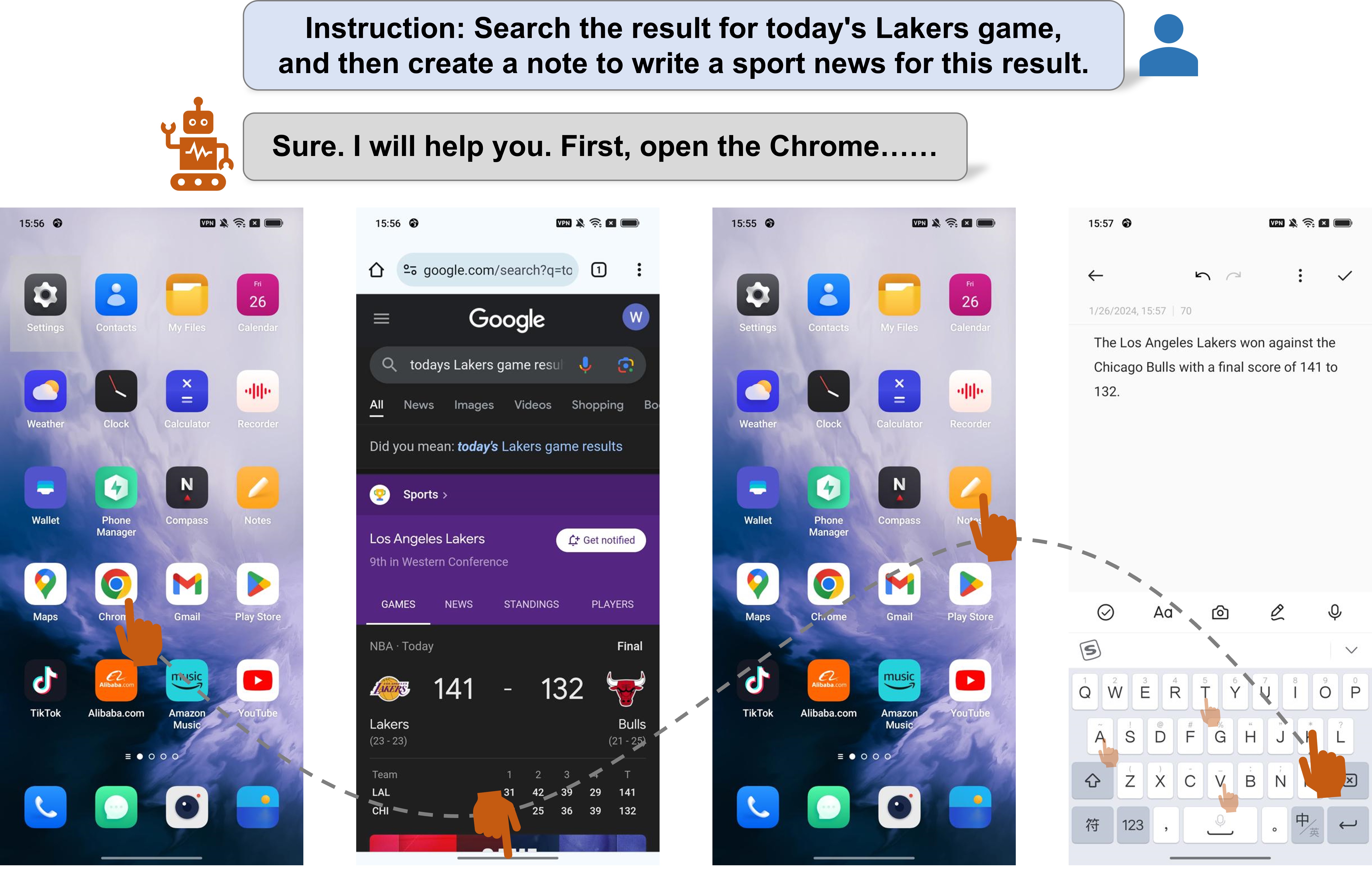}
    \caption{Mobile-Agent is an autonomous agent for operating the mobile device. Based on user instruction, Mobile-Agent can plan a series of operations to complete the requirements.}
    \label{fig:first_image}
\end{figure*}

\section{Introduction}

LLM-based agents \cite{li2023modelscope,liu2023controlllm,liu2023internchat,liu2023llava,shen2023hugginggpt,wu2023visual,yang2023gpt4tools,shen2024small,yang2023mm,hong2023metagpt,yang2023auto}, utilizing a variety of tools, have demonstrated strong capabilities in task planning and reasoning. As Multimodal Large Language Models (MLLM) \cite{liu2023visual,zhu2023minigpt,ye2023mplug,dai2023instructblip,liu2023improved,chen2023minigpt,ye2023mplugowl2,bai2023qwen,lin2023vila} rapidly progress and exhibit remarkably visual comprehension capabilities, the realization of MLLM-based agents has become feasible, also sparking the potential for a variety of innovative applications.

Recently, mobile device agent has emerged as a novel and popular application of MLLM-based agents. The agent needs to operate the mobile device based on the screen and user instructions. This requires the agent to possess both visual perception and semantic understanding capabilities. However, existing MLLMs, including the state-of-the-art GPT-4V, still lack sufficient visual perception abilities to serve as an effective agent. \cite{zheng2024gpt} points out that although GPT-4V can generate effective operations, it struggles to accurately locate the positions of these operations on the screen. This limitation hinders the ability to operations on mobile device solely through advanced MLLMs.

To address this issue, existing works have attempted to assist GPT-4V in localization by leveraging user interface layout files. \cite{yang2023appagent} extracted actionable positions on the screen by accessing Android application XML files. \cite{zheng2024gpt} used HTML code from web applications to aid in localization. These methods rely on the accessibility of underlying files. However, in many scenarios, permissions to access these files may not be available, rendering these methods ineffective. 

In order to eliminate the dependency on the underlying files in existing localization methods, in this work, we propose Mobile-Agent, an autonomous mobile device agent with visual perception. Mobile-Agent, through the visual perception module, can accurately locate operations using only screenshots from the mobile device. The visual perception module consists of detection and OCR models, responsible for describing the content of localized screen regions and identifying text within the screen, respectively. Through carefully crafted prompts, we facilitate effective interaction between the agent and tools, enabling the automation of mobile device operations. Leveraging the robust contextual capabilities of GPT-4V, Mobile-Agent achieves a self-planning capability to plan tasks holistically based on the screenshot, user instruction, and operation history. To enhance the agent's ability to identify erroneous operations and incomplete instructions, we introduce a self-reflection method. Guided by prompts, the agent continually reflects on invalid and incorrect operations, and the agent can halt once the instruction is completed. In order to comprehensively assess Mobile-Agent's capabilities, we have introduced Mobile-Eval, a benchmark centered around current mainstream mobile Apps. Mobile-Eval includes instructions for various difficulty levels. We have conducted an analysis of Mobile-Agent based on Mobile-Eval, showcasing and analyzing some of the cases within it. The experimental results indicate that Mobile-Agent exhibits remarkable instruction completion rates and operation accuracy. Even in challenging instructions, such as operating multiple Apps, Mobile-Agent is able to successfully complete the tasks.

The contributions summarized are as follows:
\begin{itemize}
    \item We propose Mobile-Agent, an autonomous mobile device agent. Mobile-Agent utilizes visual perception tools for operation localization. It can self-plan each step and complete self-reflection. Mobile-Agent solely relies on device screenshots without any system code, which is a purely vision-based solution.
    \item We introduce Mobile-Eval, a benchmark designed to assess mobile device agents. This benchmark comprises 10 commonly used Apps and features instructions with varying three difficulty levels.
    \item We conducted a comprehensive analysis of Mobile-Agent based on Mobile-Eval. We presented typical selected cases to analyze the capabilities of it.
\end{itemize}

\begin{figure*}[t]
    \centering
    \includegraphics[width=0.8\textwidth]{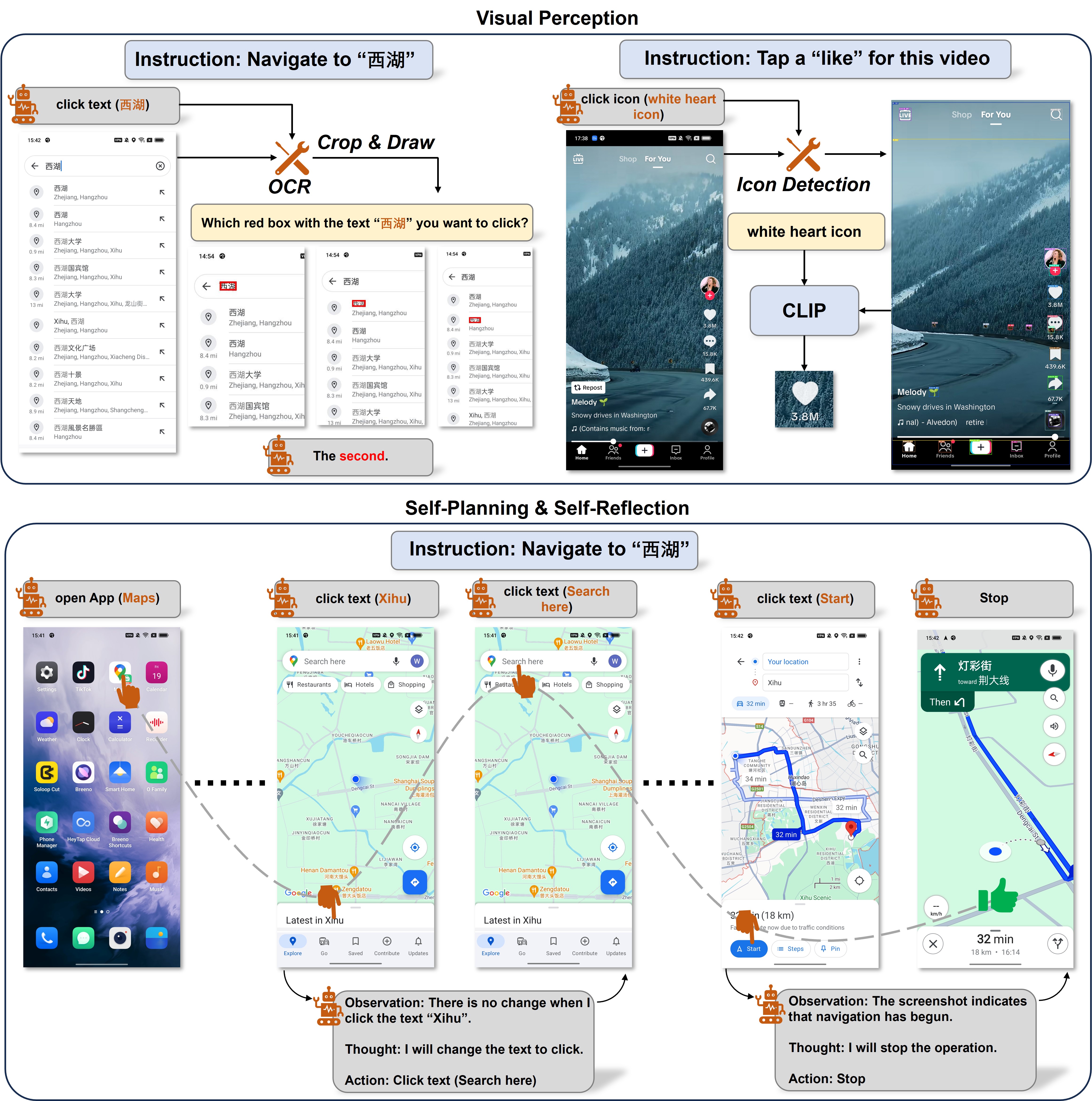}
    \caption{The framework of Mobile-Agent.}
    \label{fig:illu}
\end{figure*}

\section{Mobile-Agent}

This section introduces our Mobile-Agent framework. The framework consists of state-of-the-art MLLM GPT-4V, a text detection module for text localization, and an icon detection module for icon localization. We will first explain how to use visual tools to position the instructions generated by GPT-4V to specific locations on the mobile device. Subsequently, we will describe the workflow of the Mobile-Agent.

\subsection{Visual Perception}

\noindent \textbf{GPT-4V Lacks Localization Capability.} While GPT-4V can provide correct operations for instruction and screenshot, existing work~\cite{zheng2024gpt} indicates that GPT-4V is unable to effectively output the location where the operations take place. Therefore, we need external tools to assist GPT-4V in operation localization, allowing the operations to be output onto the mobile device screen.

\noindent \textbf{Text Localization.} When the agent needs to tap on specific text on the screen, we use an OCR tool to detect the position of the corresponding text on the screen. We will discuss three scenarios:
\begin{itemize}
    \item When the OCR detection results do not include the specified text, the agent will be instructed to either reselect the text for tapping or choose an alternative operation. This situation often occurs in complex scenarios where GPT-4V may have a small number of hallucinations.
    \item When the OCR detection results only have one instance of the specified text, we directly generate an operation to click on the center coordinates of that text box.
    \item When the OCR detection results include multiple instances of the specified text, we assess the number of results. If there are many instances, it indicates that there is too much similar content on the current screen, making it challenging for the agent to make a selection. In such cases, the agent is requested to reselect the text for tapping. If there are few instances, we crop these regions and draw detection boxes on them. Then, we use these regions to let the agent choose which one to click. When cropping, we extend the text detection boxes outward by a certain range and then draw the detection boxes on these cropped images. This is done to preserve more information and facilitate the agent's decision-making process. This process is shown in the top-left of Figure~\ref{fig:illu}.
\end{itemize}

\noindent \textbf{Icon Localization.} When the agent needs to click an icon, we use an icon detection tool and CLIP~\cite{radford2021learning} to locate the position of it. Specifically, we first request the agent to provide the attributes of the icon to click, including color and shape. Subsequently, we use Grounding DINO~\cite{liu2023grounding} with the prompt ``icon'' to identify all the icons on the screenshot. Finally, employing CLIP, we calculate the similarity between all detected icons and the description of the click region, selecting the region with the highest similarity for a click. This process is shown in the top-right of Figure~\ref{fig:illu}.

\subsection{Instruction Execution}

\noindent \textbf{Operation.} In order to better translate the actions output by the agent into operations on the screen, we define 8 operations for the Mobile-Agent:
\begin{itemize}
    \item Open App (\textit{App}): Open a specific App on the desktop page.
    \item Click the text (\textit{Text}): Click the area of the screen where the text ``\textit{Text}'' is located.
    \item Click the icon (\textit{Icon}, \textit{Position}): Click the area described by ``\textit{Icon}'' in the ``\textit{Position}''. ``\textit{Icon}'' provides a description, including attributes such as color, icon shape, etc., of the tapping location. ``\textit{Position}'' needs to be selected from top, bottom, left, right, or center, with one or two options, to minimize the possibility of errors.
    \item Type (\textit{Text}): Type the "\textit{Text}" into the current input box.
    \item Page up \& down: Used for scrolling up and down the current page.
    \item Back: Return to the last page.
    \item Exit: Return directly to the desktop from the current page.
    \item Stop: When the instruction is completed, end the entire process. 
\end{itemize}

\noindent \textbf{Self-Planning.} The Mobile-Agent completes each step of the operation iteratively. Before the iteration begins, the user needs to input an instruction. We generate the system prompt for the entire process based on the instruction. At the start of each iteration, we capture a screenshot of the current mobile screen and provide it to the agent. The agent, by observing the system prompt, operation history, and the current screen capture, outputs the next step of the operation. If the agent's output is to end the process, the iteration stops; otherwise, a new iteration continues. Mobile-Agent, utilizing the operation history, is aware of the current task progress and, based on the system prompt, generates operation on the current screenshot, thereby achieving an iterative self-planning process. This process is shown at the bottom of Figure~\ref{fig:illu}.

\noindent \textbf{Self-Reflection.} During the iteration, the agent may encounter errors, leading to the inability to complete the instruction. To improve the success rate of instruction, we have introduced a self-reflection method. This method will take effect in two situations. The first is when the agent generates an incorrect or invalid operation, causing the process to be stuck. When the agent notices that the screenshot has not changed after a particular operation, or the screenshot shows a wrong page, we will instruct the agent to try alternative operations or modify the parameters of the current operation. The second is when the agent may overlook certain requirements of complex instruction. After the agent completes all operations through self-planning, we will instruct the agent to analyze the operations, history, the current screenshot, and user instruction to determine if the instruction have been completed. If not, the agent needs to continue generating operations through self-planning. This process is shown at the bottom of Figure~\ref{fig:illu}.

\noindent \textbf{Prompt Format.} To better implement the functionalities described above, we drew inspiration from the prompt format used by ReAct. We require the agent to output three components: Observation, Thought, and Action. Observation is a description by the agent of the current screenshot and the history of operations. This helps the agent to notice updates in the screenshot and promptly identify errors based on historical records. Thought represents the agent's consideration of the next step of operation generated from the Observation and the instruction. The agent needs to describe the upcoming operation in the Thought. Action requires the agent to choose one of eight operations and parameters based on Thought.

\section{Experiments}

In this section, we will conduct a comprehensive evaluation of the Mobile-Agent. We use the Android operating system due to its convenient operation invocation interfaces. We will explore other operating systems in future work. Our experiments are primarily divided into two parts: quantitative experiments and qualitative experiments. In the quantitative experiments, we will evaluate the Mobile-Agent on our proposed Mobile-Eval benchmark. In the qualitative experiments, we will analyze specific cases.

\begin{table*}[t]
    \centering
    \renewcommand{\arraystretch}{1.1}
    \setlength{\tabcolsep}{10pt}
    \scalebox{0.75}{
    \begin{tabular}{p{2cm} | p{18cm} }
    \hline
    \toprule 
     \textbf{Application}&\textbf{Instruction}\\
     \midrule
     \multirow{3}{*}{Alibaba.com}&Help me find caps in Alibaba.com.\\
     &Help me find caps in Alibaba.com. If the "Add to cart" is available in the item information page, please add the item to my cart.\\
     &I want to buy a cap. I've heard things are cheap on Alibaba.com. Maybe you can find it for me.\\
     \midrule
     \multirow{3}{*}{Amazon Music}&Search singer Jay Chou in Amazon Music.\\
     &Search a music about "agent" in Amazon Music and play it.\\
     &I want to listen music to relax. Find an App to help me.\\
     \midrule
     \multirow{3}{*}{Chrome}&Search result for today's Lakers game.\\
     &Search the information about Taylor Swift.\\
     &I want to know the result for today's Lakers game. Find an App to help me.\\
     \midrule
     \multirow{3}{*}{Gmail}&Send an empty email to to \{address\}.\\
     &Send an email to \{address\} to tell my new work.\\
     &I want to let my friend know my new work, and his address is \{address\}. Find an App to help me.\\
     \midrule
     \multirow{3}{*}{Google Maps}&Navigate to Hangzhou West Lake.\\
     &Navigate to a nearby gas station.\\
     &I want to go to Hangzhou West Lake, but I don't know the way. Find an App to help me.\\
     \midrule
     \multirow{3}{*}{Google Play}&Download WhatsApp in Play Store.\\
     &Download Instagram in Play Store.\\
     &I want WhatsApp on my phone. Find an App to help me.\\
     \midrule
     \multirow{3}{*}{Notes}&Create a new note in Notes.\\
     &Create a new note in Notes and write "Hello, this is a note", then save it.\\
     &I suddenly have something to record, so help me find an App and write down the following content: meeting at 3pm.\\
     \midrule
     \multirow{3}{*}{Settings}&Turn on the dark mode.\\
     &Turn on the airplane mode.\\
     &I want to see the real time internet speed at the battery level, please turn on this setting for me.\\
     \midrule
     \multirow{3}{*}{TikTok}&Swipe a video about pet cat in TikTok and click a "like" for this video.\\
     &Swipe a video about pet cat in TikTok and comment "Ohhhh, so cute cat!".\\
     &Swipe videos in TikTok. Click "like" for 3 pet video cat.\\
     \midrule
     \multirow{3}{*}{YouTube}&Search for videos about Stephen Curry on YouTube.\\
     &Search for videos about Stephen Curry on YouTube and open "Comments" to comment "Oh, chef, your basketball spirit has always inspired me".\\
     &I need you to help me show my love for Stephen Curry on YouTube.\\
     \midrule
     \multirow{3}{*}{Multi-App}&Open the calendar and look at today's date, then go to Notes and create a new note to write "Today is [today's data]".\\
     &Check the temperature in the next 5 days, and then create a new note in Notes and write a temperature analysis.\\
     &Search the result for today's Lakers game, and then create a note in Notes to write a sport news for this result.\\
    \bottomrule 
    \hline
  \end{tabular}}
    \caption{The applications and instructions used in Mobile-Eval.}
    \label{tab:mobile-eval}
\end{table*} 

\subsection{Setup}

\noindent \textbf{Mobile-Eval.} To comprehensively evaluate the capabilities of Mobile-Agent, we introduce Mobile-Eval, a benchmark based on current mainstream Apps. Mobile-Eval consists of a total of 10 commonly used Apps on mobile devices. To assess the multi-application usage capability of the agent, we have also introduced instructions that require the simultaneous use of two Apps. We designed three instructions for each App. The first instruction is relatively simple, requiring only the completion of basic App operations. The second instruction adds some additional requirements to the first one, making it more challenging. The third instruction involves abstract user instruction, where the user does not explicitly specify which App to use or what operation to perform, leaving the agent to make its own judgment. In Table~\ref{tab:mobile-eval}, we present the Apps and instructions used in Mobile-Eval.

\noindent \textbf{Metrics.} We have designed four metrics to assess the performance of the Mobile-Agent from different perspectives:
\begin{itemize}
    \item Success (Su): If the Mobile-Agent completes the instruction, it is considered successful.
    \item Process Score (PS): This metric measures the accuracy of each step in the execution of instructions. Specifically, it equals the number of correct steps divided by the total number of steps. Although the agent may not ultimately succeed in some instructions, each correct step contributes to the Planning Score. 
    \item Relative Efficiency (RE). We manually performed each instruction and recorded the number of steps taken by a human. We consider human operation as the optimal solution. We will compare the number of steps taken by Mobile-Agent with the steps taken by humans to demonstrate whether Mobile-Agent can use the mobile device more efficiently.
    \item Completion Rate (CR). We calculate the number of human-operated steps that Mobile-Agent is able to complete, divided by the total number of steps taken by a human, to demonstrate the completion rate of Mobile-Agent for a given instruction. If the instruction is completed, this metric will be equal to 1. 
\end{itemize}

\subsection{Quantitative Results}

\begin{table*}[t]
    \centering
    \renewcommand{\arraystretch}{1.2}
    \setlength{\tabcolsep}{7pt}
    \scalebox{0.9}{
    \begin{tabular}{l | c c c c | c c c c | c c c c}
    \hline
    \toprule
    \multicolumn{1}{l}{\multirow{2}{*}{\textbf{App}}}&\multicolumn{4}{c}{\textsc{Instruction 1}}&\multicolumn{4}{c}{\textsc{Instruction 2}}&\multicolumn{4}{c}{\textsc{Instruction 3}}\\
    \cmidrule(lr){2-5}
    \cmidrule(lr){6-9}
    \cmidrule(lr){10-13}
    &SU&PS&RE&CR&SU&PS&RE&CR&SU&PS&RE&CR\\
    \midrule
    Alibaba.com&\color{darkgreen}\Checkmark&0.75&4 / 3&100\%&\color{red}\ding{55}&0.39&13 / 8&62.5\%&\color{darkgreen}\Checkmark&0.9&10 / 9&100\%\\
    Amazon Music&\color{red}\ding{55}&0.44&9 / 5&80.0\%&\color{darkgreen}\Checkmark&0.75&8 / 6& 100\%&\color{red}\ding{55}&0.50&12 / 3& 66.7\%\\
    Chrome&\color{darkgreen}\Checkmark&1.00&4 / 4&100\%&\color{darkgreen}\Checkmark&0.8&5 / 4& 100\%&\color{darkgreen}\Checkmark&0.43&8 / 5&100\%\\
    Gmail&\color{darkgreen}\Checkmark&1.00&4 
/ 4&100\%&\color{red}\ding{55}&0.56&9 / 8&37.5\%&\color{red}\ding{55}&0.56&9 / 8&37.5\%\\
    Google Maps&\color{darkgreen}\Checkmark&1.00&5 / 5&100\%&\color{darkgreen}\Checkmark&1.00&6 / 6&100\%&\color{darkgreen}\Checkmark&1.00&6 / 6&100\%\\
    Google Play&\color{darkgreen}\Checkmark&1.00&3 / 3&100\%&\color{darkgreen}\Checkmark&0.50&10 / 4&100\%&\color{darkgreen}\Checkmark&1.00&3 / 3&100\%\\
    Notes&\color{darkgreen}\Checkmark&0.57&7 / 4&100\%&\color{darkgreen}\Checkmark&0.67&6 / 4&100\%&\color{darkgreen}\Checkmark&1.00&5 / 5&100\%\\
    Settings&\color{darkgreen}\Checkmark&1.00&4 / 4&100\%&\color{darkgreen}\Checkmark&1.00&4 / 4&100\%&\color{darkgreen}\Checkmark&1.00&5 / 5&100\%\\
    TikTok&\color{darkgreen}\Checkmark&1.00&4 / 4&100\%&\color{darkgreen}\Checkmark&1.00&10 / 10&100\%&\color{darkgreen}\Checkmark&1.00&7 / 7&100\%\\
    YouTube&\color{darkgreen}\Checkmark&1.00&4 / 4&100\%&\color{darkgreen}\Checkmark&1.00&9 / 9&100\%&\color{darkgreen}\Checkmark&1.00&7 / 7&100\%\\
    Multi-App&\color{darkgreen}\Checkmark&1.00&6 / 6&100\%&\color{darkgreen}\Checkmark&1.00&6 / 6&100\%&\color{darkgreen}\Checkmark&1.00&10 / 10&100\%\\
    \midrule
    Avg.&0.91&0.89&4.9 / 4.2&98.2\%&0.82&0.77&7.9 / 6.3&90.9\%&0.82&0.84&7.5 / 6.2&91.3\%\\
    \bottomrule
    \hline
    \end{tabular}
    }
    \caption{The overall evaluation results of Mobile-Agent on Mobile-Eval, where the two values of RE represent the number of steps taken by Mobile-Agent and human, respectively.}
    \label{tab:results}
\end{table*} 

We present the experimental results in Table~\ref{tab:results}. Firstly, across the three instructions, Mobile-Agent achieved completion rates of 91\%, 82\%, and 82\% respectively. Despite some instructions not being successfully executed, the completion rates for all three types of instructions exceeded 90\%. Next, we can observe from the PS metric that Mobile-Agent has a high probability of producing correct operations across the three instructions, achieving around 80\%. Finally, the RE metric indicates that Mobile-Agent can achieve an 80\% capability of reaching human-optimal operations. The above results collectively indicate the effectiveness of Mobile-Agent as a mobile device assistant.

It is worth noting that the PS values for some instructions don't reach 1, indicating that Mobile-Agent may make some invalid or incorrect operations. However, in these cases, most instructions were ultimately completed. This suggests that Mobile-Agent possesses good self-reflective capabilities. Even in the presence of invalid or incorrect operations, it can reflect based on the screenshots and ultimately correct its mistakes. This is crucial for mobile device agents because, like humans, they cannot guarantee that all operations are correct, and agents must have the ability to correct errors.

\subsection{Case Study}

\begin{figure*}[t]
    \centering
    \includegraphics[width=0.85\textwidth]{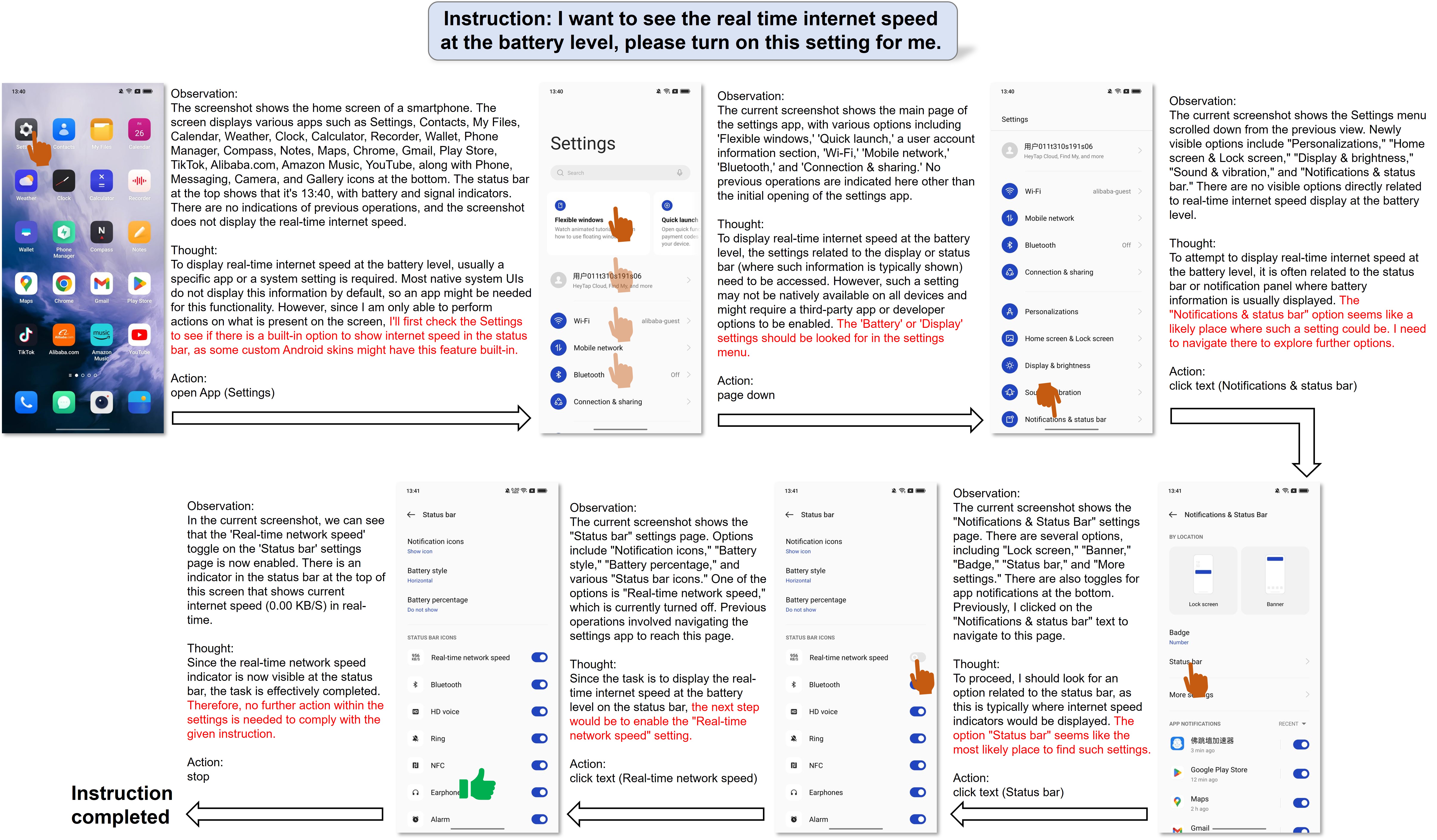}
    \caption{Case of instruction comprehension and execution planning.}
    \label{fig:case1}
\end{figure*}

\begin{figure*}[t]
    \centering
    \includegraphics[width=0.75\textwidth]{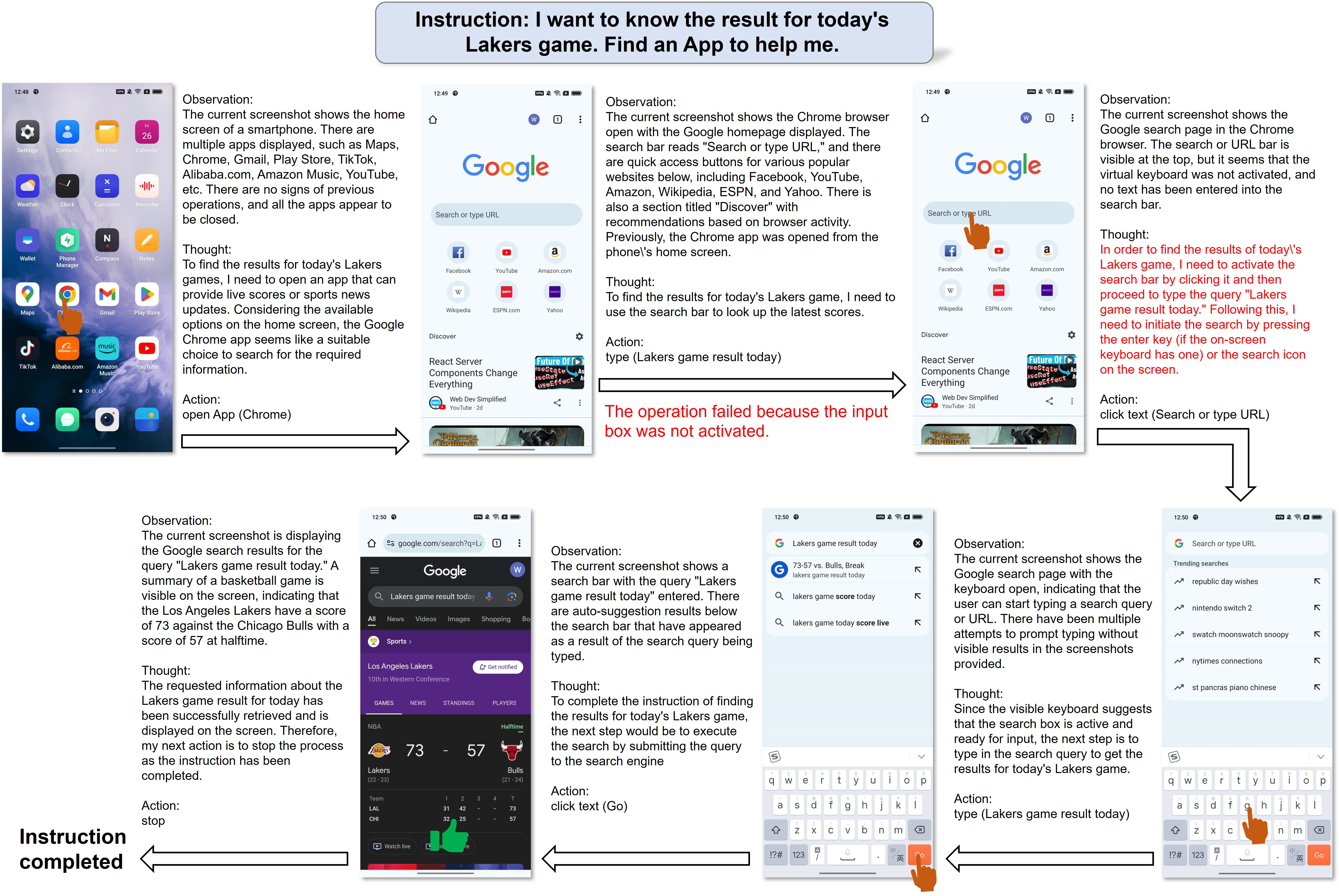}
    \caption{Case of self-reflection and error correction after using invalid operations.}
    \label{fig:case2}
\end{figure*}

\begin{figure*}[t]
    \centering
    \includegraphics[width=0.71\textwidth]{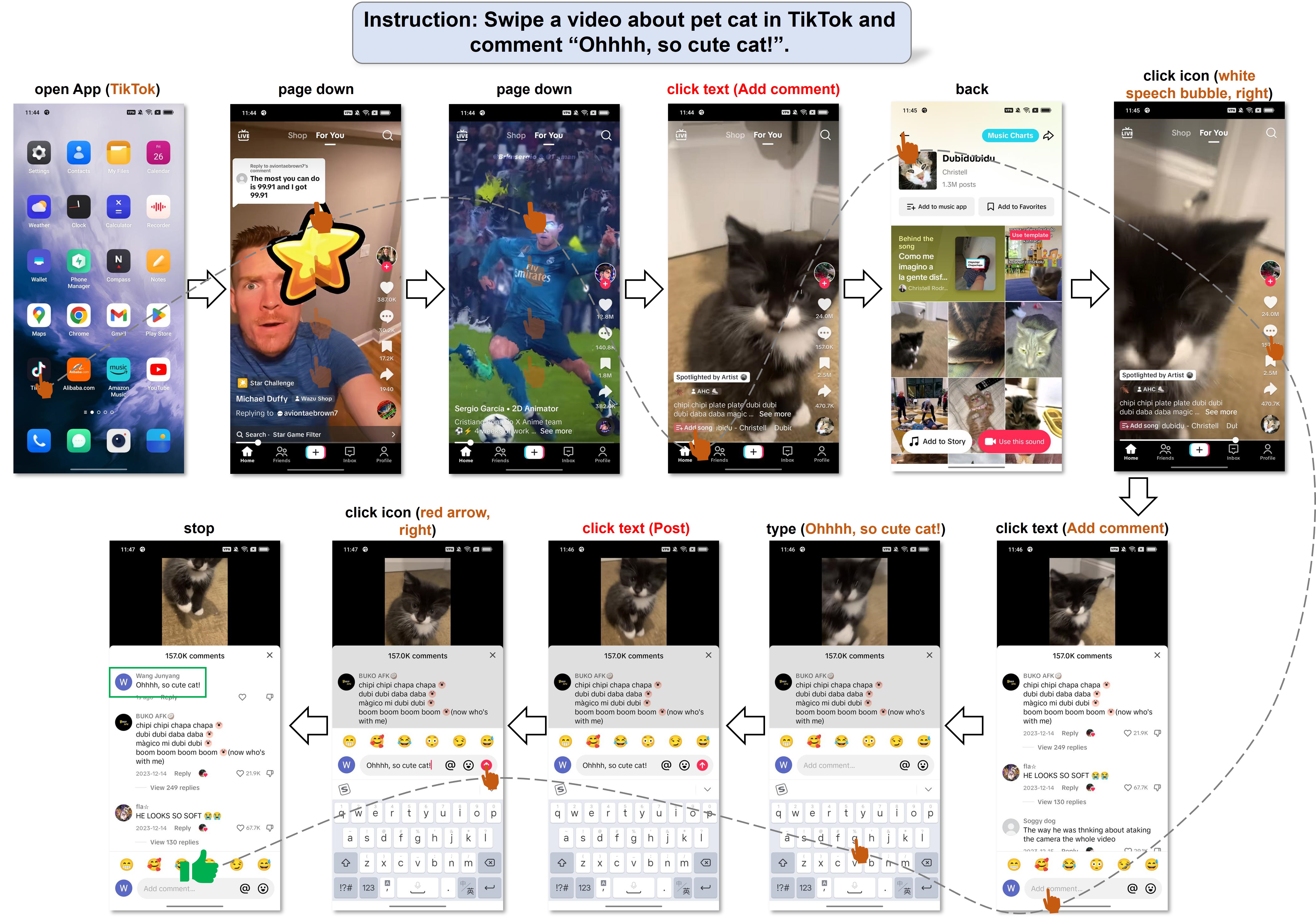}
    \caption{Case of self-reflection and error correction after using invalid and incorrect operations, where the operation ``click text (Add comment)'' leads to an incorrect page and the operation ``click text (Post)'' is an invalid operation. The invalid and incorrect operations are highlighted in red font.}
    \label{fig:case7}
\end{figure*}

\begin{figure*}[t]
    \centering
    \includegraphics[width=0.68\textwidth]{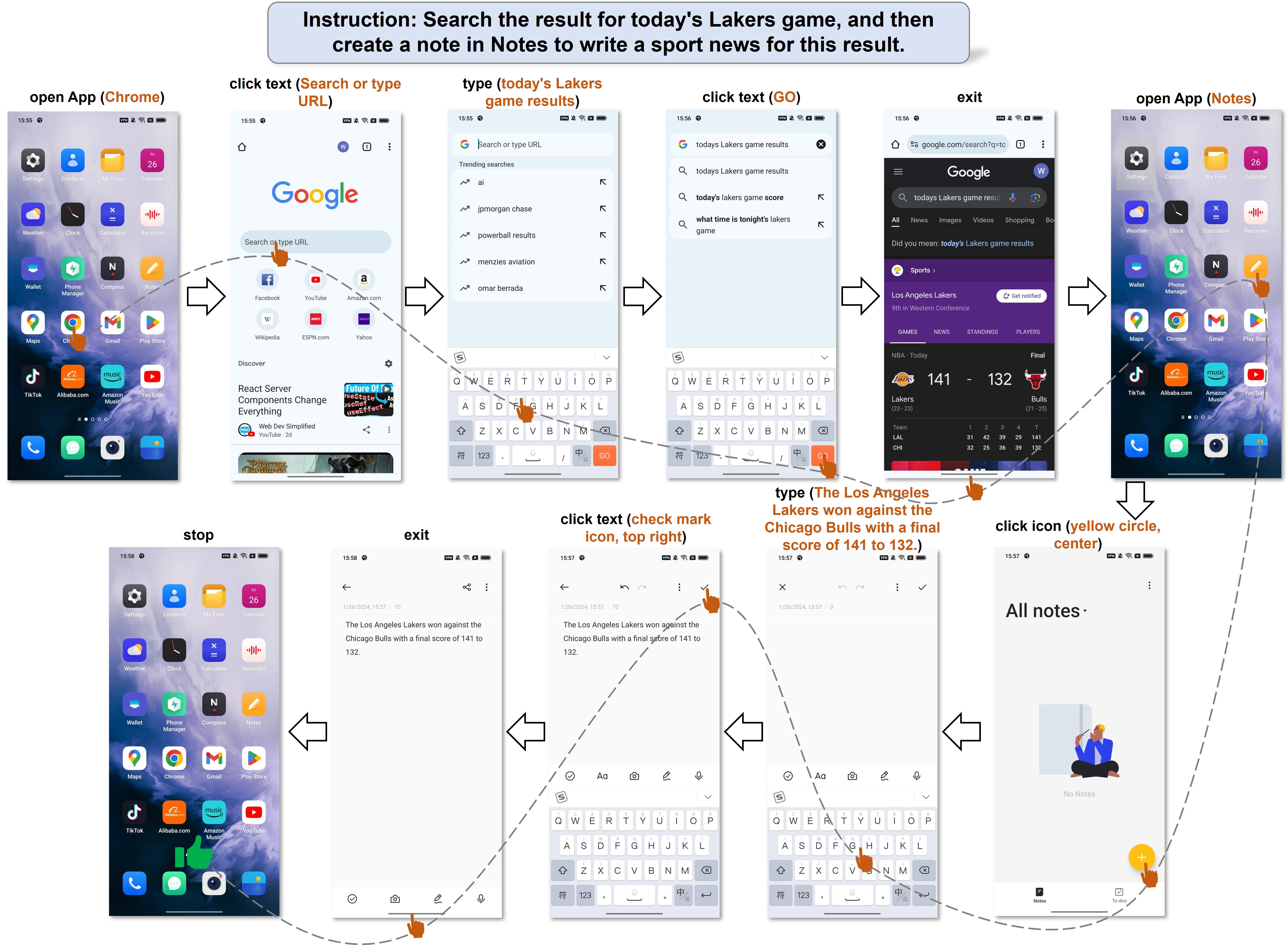}
    \caption{Case of operating multiple Apps to search game result.}
    \label{fig:case9}
\end{figure*}

\begin{figure*}[t]
    \centering
    \includegraphics[width=0.45\textwidth]{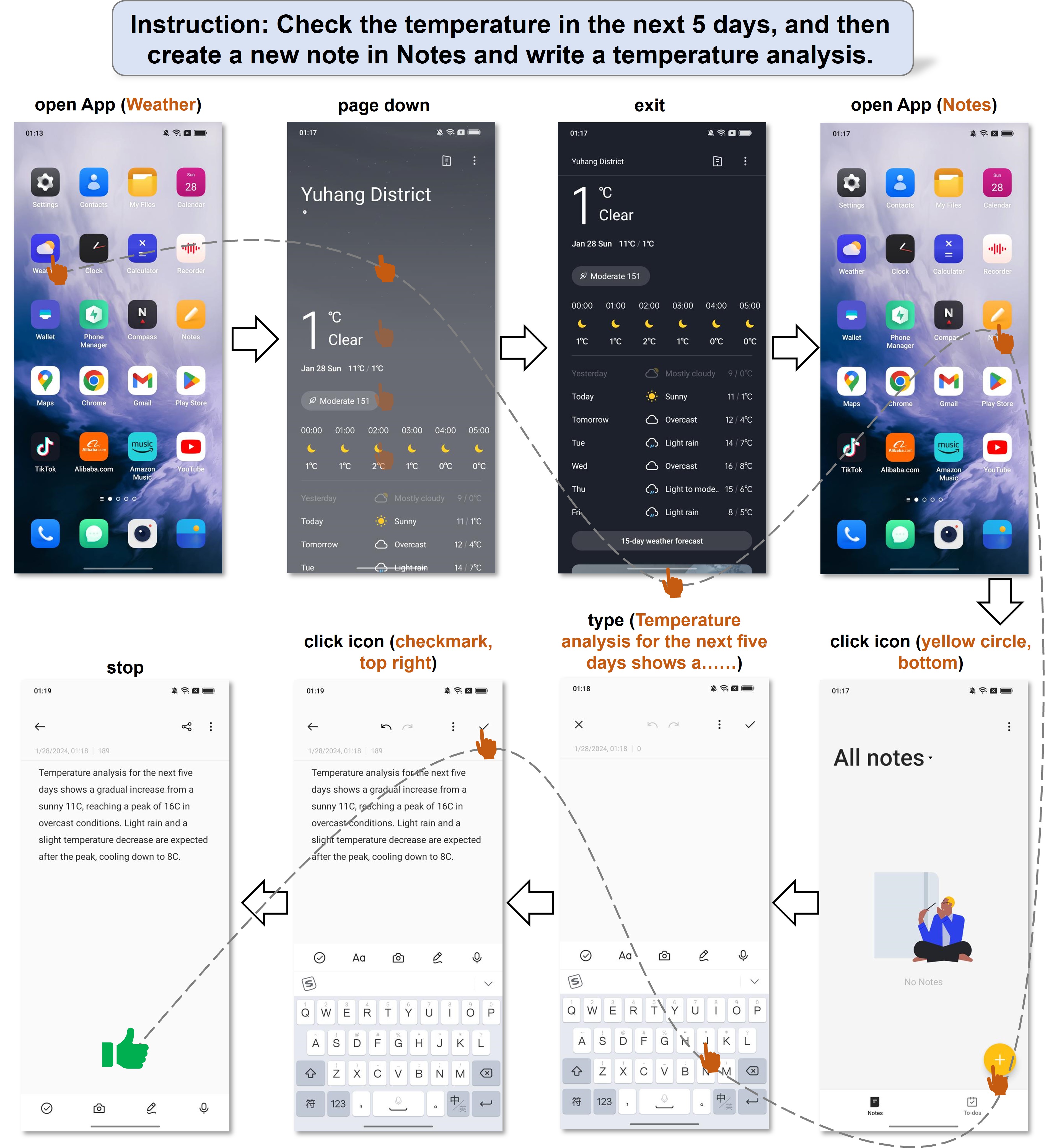}
    \caption{Case of operating multiple Apps to write a temperature analysis.}
    \label{fig:case12}
\end{figure*}

\begin{figure*}[t]
    \centering
    \includegraphics[width=0.45\textwidth]{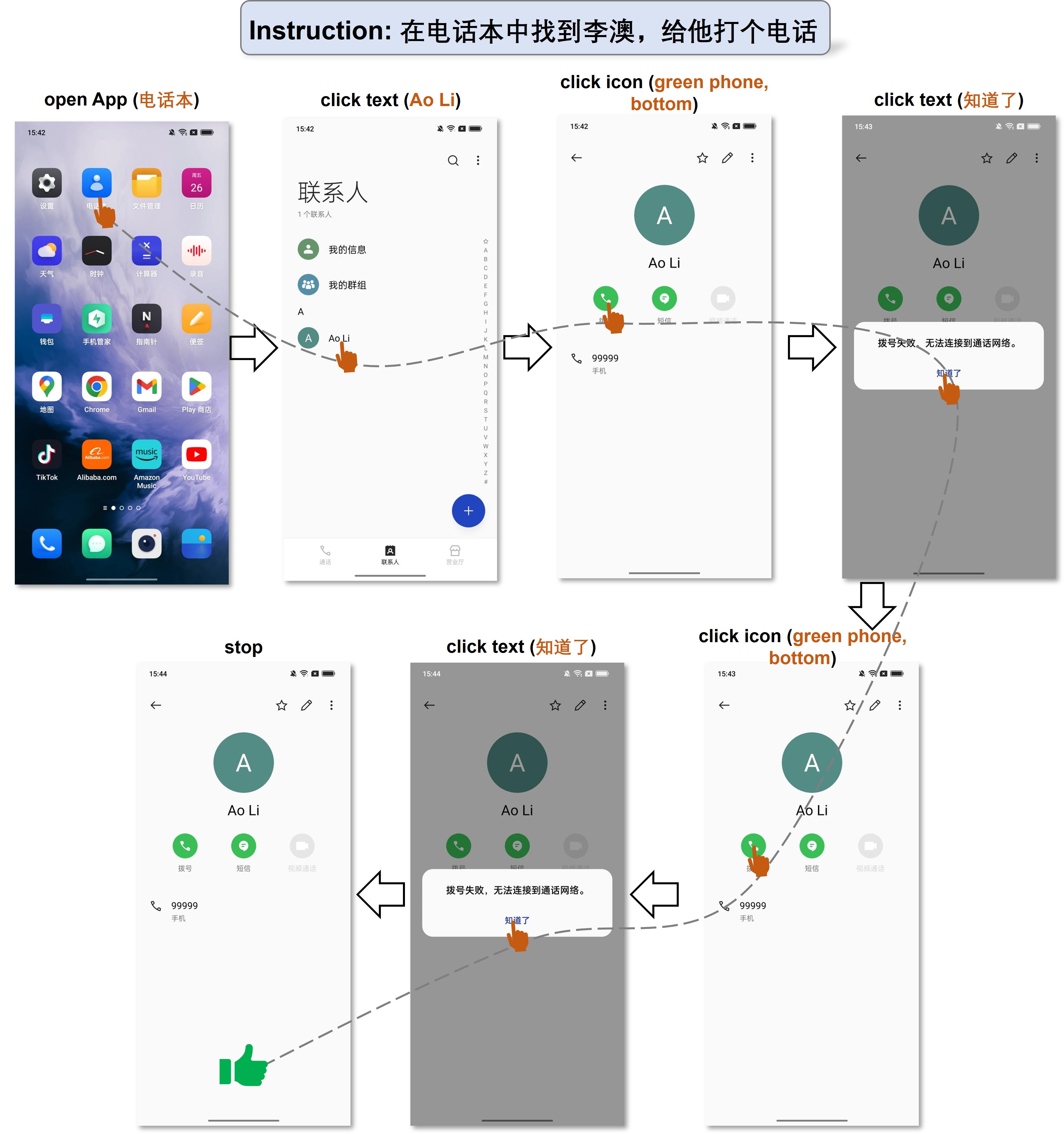}
    \caption{Case of operating Chinese system and App.}
    \label{fig:case10}
\end{figure*}

\begin{figure*}[t]
    \centering
    \includegraphics[width=0.45\textwidth]{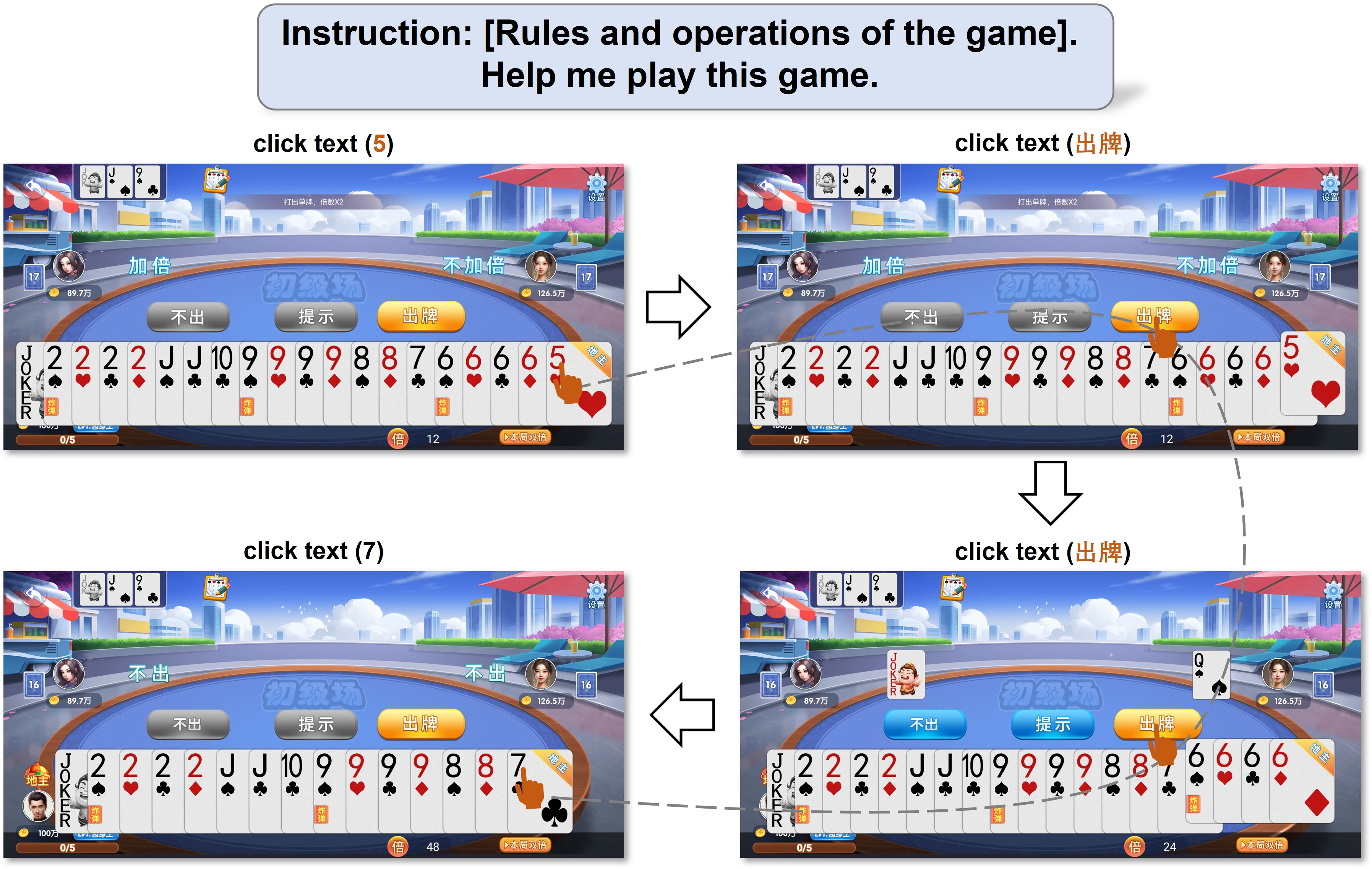}
    \caption{Case of playing games.}
    \label{fig:case11}
\end{figure*}

\begin{figure*}[t]
    \centering
    \includegraphics[width=0.7\textwidth]{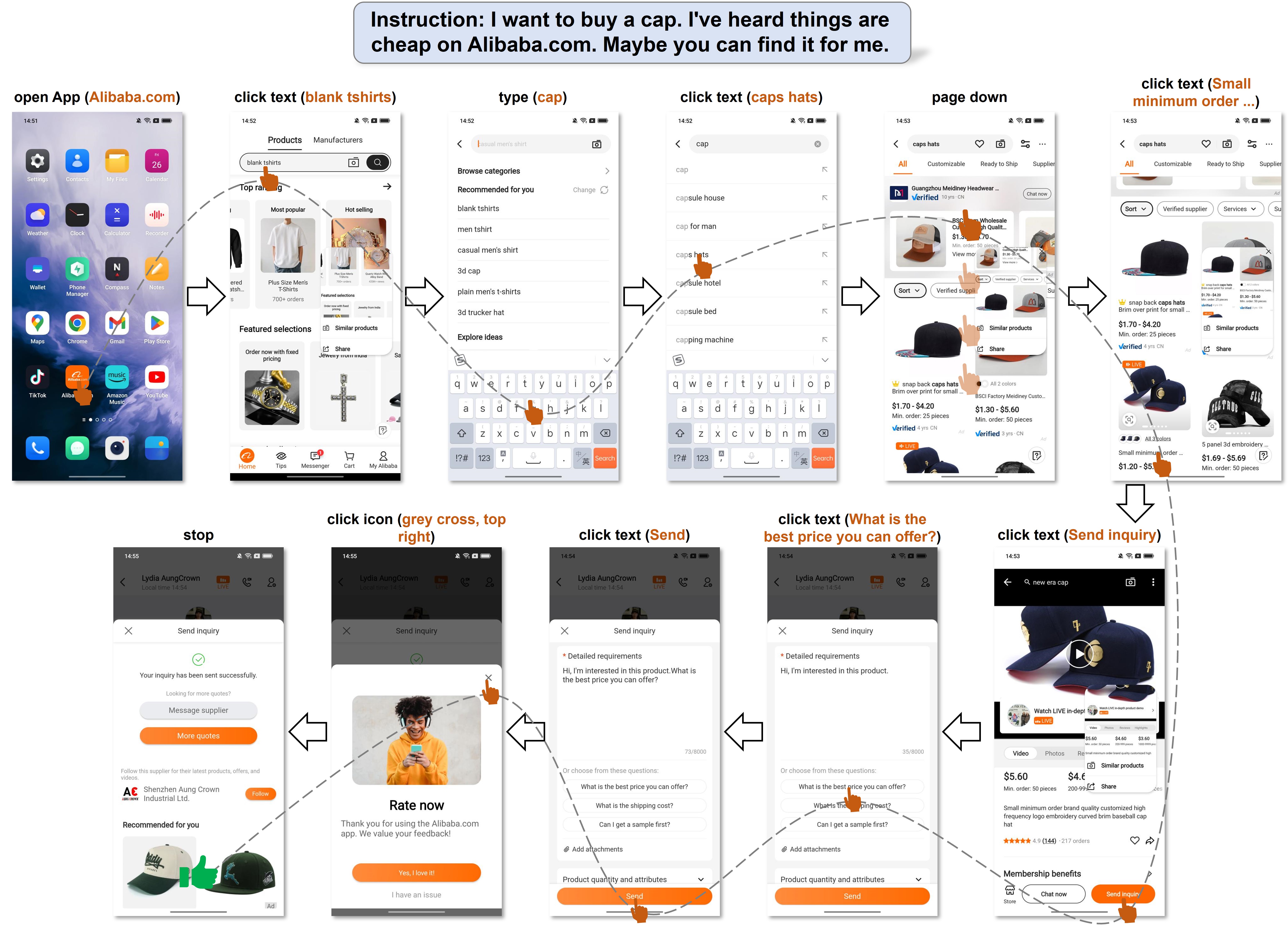}
    \caption{Case of wholesale caps from Alibaba.com.}
    \label{fig:case3}
\end{figure*}

\begin{figure*}[!ht]
    \centering
    \includegraphics[width=0.52\textwidth]{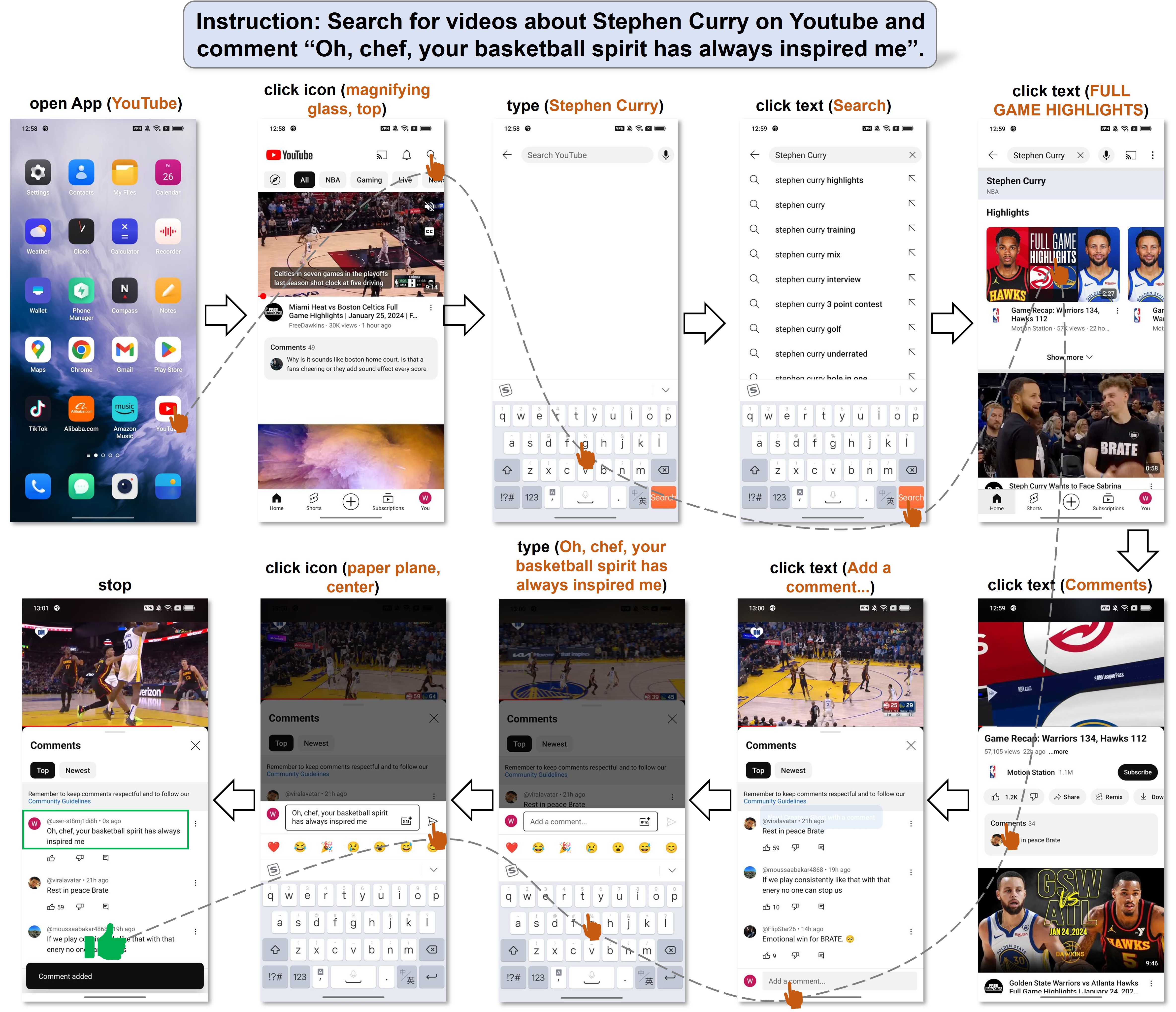}
    \caption{Case of searching video from YouTube and commenting this video.}
    \label{fig:case8}
\end{figure*}

\begin{figure*}[!ht]
    \centering
    \includegraphics[width=0.47\textwidth]{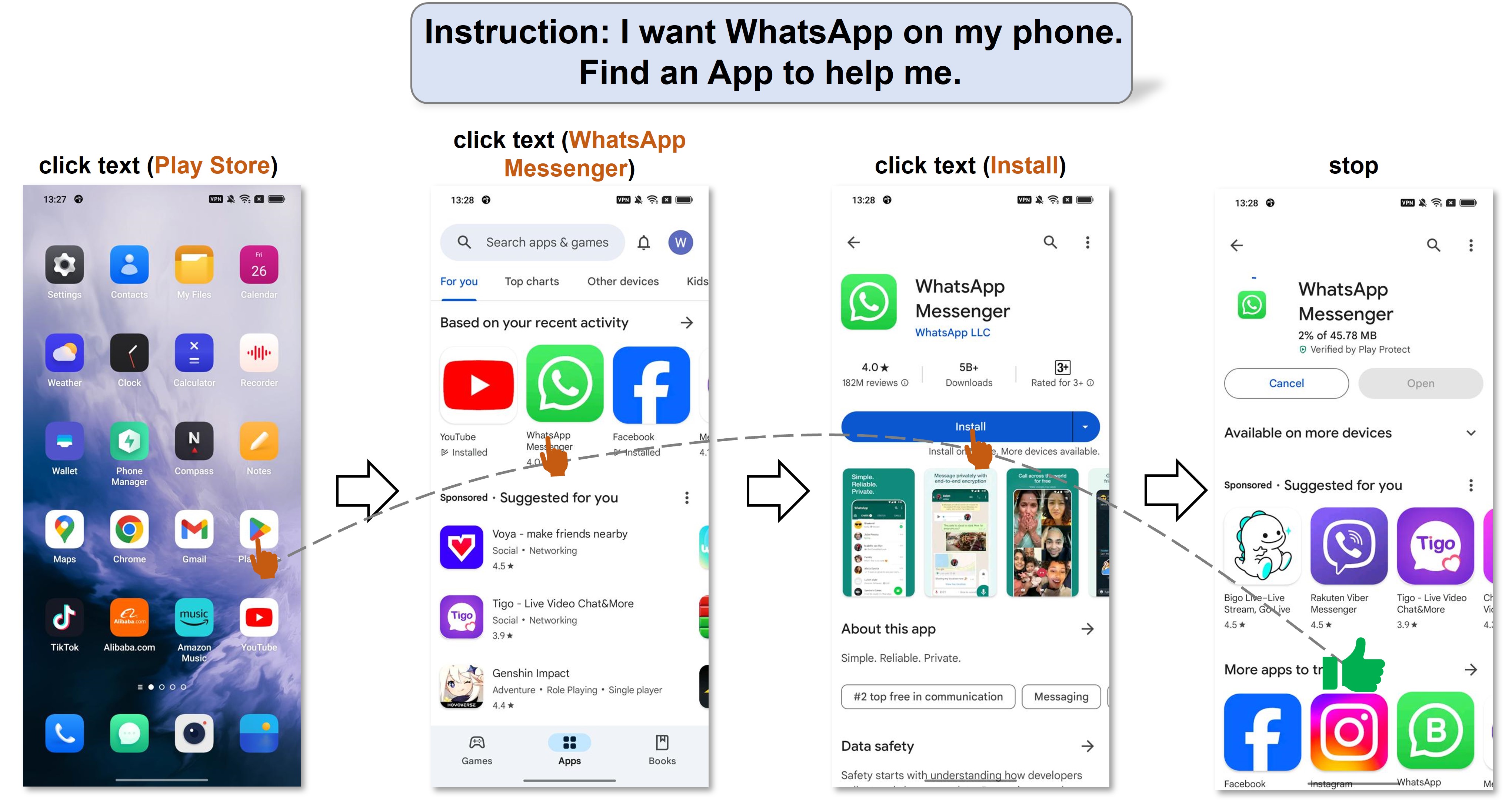}
    \caption{Case of downloading specific App in Google Play.}
    \label{fig:case5}
\end{figure*}

\begin{figure*}[!ht]
    \centering
    \includegraphics[width=0.44\textwidth]{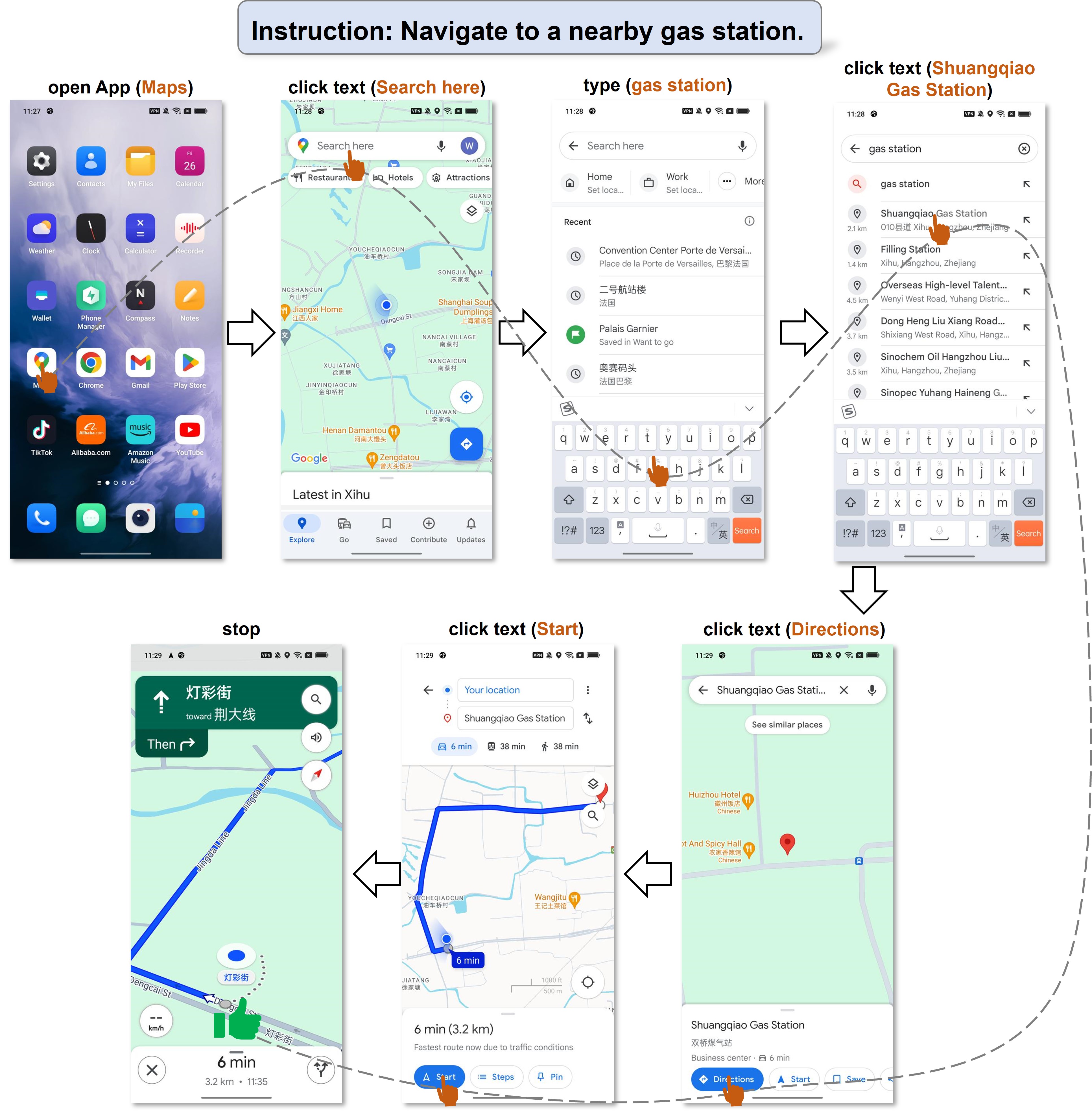}
    \caption{Case of using a map App for navigation.}
    \label{fig:case6}
\end{figure*}

In Figure~\ref{fig:case1}, we demonstrate Mobile-Agent's ability to understand user instructions and autonomously plan its operations. While the instructions may not contain specific operations, Mobile-Agent successfully comprehended the user's requirements and translated them into concrete executable operations. Subsequently, the agent carried out the instructions through a series of planning steps.

In Figure~\ref{fig:case2}, we showcase Mobile-Agent's ability to reflect when faced with invalid or erroneous instructions. In this case, Mobile-Agent initially used an invalid operation, resulting in no change in the screenshot. After reflection, Mobile-Agent corrected the error, continued with the operation, and ultimately completed the instruction. In Figure~\ref{fig:case7}, we show another case. Faced with two consecutive invalid or incorrect operations, Mobile-Agent was able to promptly correct the operations to ensure the smooth execution of the entire process.

In Figure~\ref{fig:case9}, \ref{fig:case12}, we showcased Mobile-Agent's capability in scenarios involving the operations across multiple apps. This requires the agent to possess a certain level of memory capacity to facilitate information transfer between the two Apps. As evident from the case, Mobile-Agent accurately conveys information from the first opened App to the second one and can generate reprocessed content.

In Figure~\ref{fig:case10}, we demonstrate the multilingual capability of Mobile-Agent. Although GPT-4V may currently have limitations in handling Chinese, its powerful visual perception allows it to handle simple Chinese scenarios effectively. In Figure~\ref{fig:case11}, we showcase the ability of Mobile-Agent to play poker games. After describing the rules of the game, Mobile-Agent can execute operations according to the specified rules.

\begin{figure*}[!ht]
    \centering
    \includegraphics[width=0.55\textwidth]{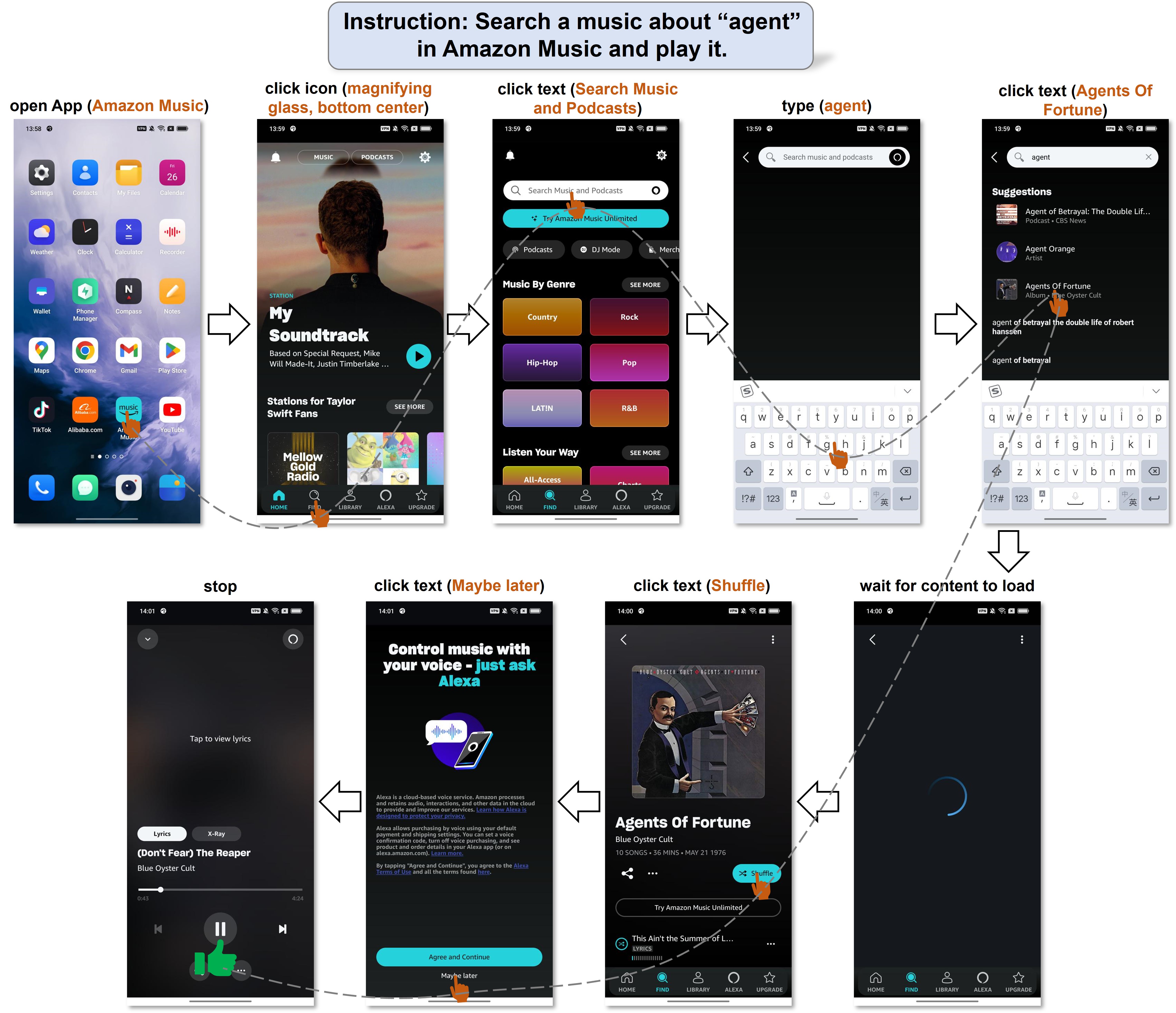}
    \caption{Case of using Amazon Music to search and play a music with specific content.}
    \label{fig:case4}
\end{figure*}

In Figure~\ref{fig:case3}, \ref{fig:case8}, \ref{fig:case5}, \ref{fig:case6}, \ref{fig:case4}, we demonstrate the powerful capabilities of Mobile-Agent on Mobile-Eval. Despite variations in the user interfaces and operations of these Apps, and the presence of challenging instructions, Mobile-Agent successfully completes the given instructions.

\section{Related Work}

\subsection{LLM-based Agent}

With the rapid advancement of Large Language Models (LLMs), agents built upon these models have notched up impressive achievements across a burgeoning spectrum of tasks~\cite{li2023modelscope,liu2023controlllm,liu2023internchat,liu2023llava,shen2023hugginggpt,wu2023visual,yang2023gpt4tools,shen2024small,yang2023mm,hong2023metagpt,yang2023auto}. Functioning as the core, these agents adeptly interpret user instructions and deploy a versatile array of tools to execute intricate tasks. The expansive integration of diverse tools liberates LLMs from the confines of pure text processing. Currently, LLM-based agents are flourishing in diverse domains, showcasing prowess in tasks such as image and video editing, image generation, visual question answering, intelligent predictions, and more. This underscores the transformative impact of LLMs on the landscape of AI applications.

\subsection{Agent for Mobile Device}

The application of agents to operate terminal devices is becoming a hotspot. AppAgent~\cite{yang2023appagent} is a mobile App assistant based on GPT-4V. They label manipulable regions of the app's UI with semi-transparent tags by invoking XML files from the Android system. The agent acquires operational capabilities through three methods: self-exploration, observing user video demos, and utilizing user documents. After a certain degree of exploration, the agent gains a sufficient understanding of operable regions, allowing it to execute correct operations based on instructions.

\section{Conclusion}

In this work, we introduce Mobile-Agent, an autonomous multi-modal agent, proficient in operating a broad spectrum of mobile applications through a unified visual perception framework. Mobile-Agent employs visual perception tools to precisely identify and locate visual and textual elements within the app's interface. Utilizing the perceived visual context, it autonomously plans, decomposes complex tasks, and navigates through mobile apps step by step. Unlike previous solutions relying on XML mobile system metadata, Mobile-Agent offers enhanced adaptability across diverse mobile operating environments in a vision-centric manner, obviating the need for system-specific customizations. Through experiments, we demonstrate the effectiveness and efficiency of Mobile-Agent across various dimensions. This showcases its potential as a versatile and adaptable solution for interacting with mobile applications in a language-agnostic manner.

\bibliographystyle{unsrtnat}
\bibliography{references}

\end{document}